\documentclass[sigconf]{acmart}

\renewcommand\footnotetextcopyrightpermission[1]{} 
\AtBeginDocument{%
  \providecommand\BibTeX{{%
    \normalfont B\kern-0.5em{\scshape i\kern-0.25em b}\kern-0.8em\TeX}}}

\setcopyright{acmcopyright}
\copyrightyear{2018}
\acmYear{2018}
\acmDOI{XXXXXXX.XXXXXXX}
\usepackage{multirow}

\begin{document}
\title{View Adaptive Light Field Deblurring Networks with Depth Perception}

\author{
 Zeqi Shen$^{1,2}$, Shuo Zhang$^{1,2,3}$, Zhuhao Zhang$^{1}$, Qihua Chen$^{1}$,Xueyao Dong$^{1}$, Youfang Lin$^{1,2,3}$}
\affiliation{
\institution{$^1$School of Computer and Information Technology, Beijing Jiaotong University, Beijing, China}
\institution{$^2$Beijing Key Laboratory of Traffic Data Analysis and Mining, Beijing, China}
\institution{$^3$CAAC Key Laboratory of Intelligent Passenger Service of Civil Aviation, Beijing, China}
\country{}
}
\email{{shenzeqi,zhangshuo,19281058,19281062,19281063,yflin}@bjtu.edu.cn}



\begin{abstract}
The Light Field (LF) deblurring task is a challenging problem as the blur images are caused by different reasons like the camera shake and the object motion. The single image deblurring method is a possible way to solve this problem. However, since it deals with each view independently and cannot effectively utilize and maintain the LF structure, the restoration effect is usually not ideal. Besides, the LF blur is more complex because the degree is affected by the views and depth. Therefore, we carefully designed a novel LF deblurring network based on the LF blur characteristics. On one hand, since the blur degree varies a lot in different views, we design a novel view adaptive spatial convolution to deblur blurred LFs, which calculates the exclusive convolution kernel for each view. On the other hand, because the blur degree also varies with the depth of the object, a depth perception view attention is designed to deblur different depth areas by selectively integrating information from different views. Besides, we introduce an angular position embedding to maintain the LF structure better, which ensures the model correctly restores the view information. Quantitative and qualitative experimental results on synthetic and real images show that the deblurring effect of our method is better than other state-of-the-art methods.
\end{abstract}



\maketitle
\section{Introduction}
\label{intro}
There are many reasons for image blur, such as camera shake and object motion. Image blur seriously affects the visual quality of images and the performance of some common computer vision tasks such as tracking~\cite{Guo-et-al:traking2,Jin-et-al:traking}, SLAM~\cite{Lee-et-al:slam}, and object detection~\cite{Zheng-et-al:detection}.
Therefore, how to remove the image blur effectively and efficiently has attracted extensive attention from researchers. In recent years, with the rapid development of deep learning, some well-designed single image deblurring algorithms~\cite{Nah-et-al:s11,Suin-et-al:s14,Kupyn-et-al:s13,Zhang-et-al:dmphn,Zamir-et-al:mprnet} have made remarkable progress. 
The single image blur can be modeled as:
\begin{equation}
    I_{B}(x,y) =\int_{P^t_{x,y}}I^t_{S}(x,y)\, dt,
\end{equation}
where $I_{B}(x,y)$ denotes the blurred image, $I^t_{S}(x,y)$ denotes the sharp image, $P^t_{x,y}$ denotes the motion path of the pixel $(x,y)$ within the exposure time $t$. The goal of the deblurring task is to recover $I_{S}$ from $I_{B}$.

With the maturity of commercial Light Field (LF) cameras, which can obtain images from different views through one shot, researchers and enterprises have paid more attention to Light Field images (LFs). 
Limited by hardware equipment, it is easy to cause image blur when shooting LFs, which limits the application scene of LFs. Therefore, the progress of LF deblurring task can promote the development of various LF applications such as virtual reality~\cite{Overbeck-et-al:vr}, salient object detection~\cite{Jing-et-al:sod}, 3D reconstruction~\cite{Li-et-al:3d-reconstruction} and depth estimation~\cite{Li-et-al:lfde3}. 

\begin{figure} [!t]
\small
\begin{minipage}{.33\linewidth}
  \centerline{Blur Image}
\end{minipage}
\begin{minipage}{.33\linewidth}
  \centerline{Sharp Image}
\end{minipage}
\begin{minipage}{.32\linewidth}
  \centerline{Ground Truth}
\end{minipage}
\begin{minipage}{.99\linewidth}
   \centerline{\includegraphics[width=.99\linewidth]{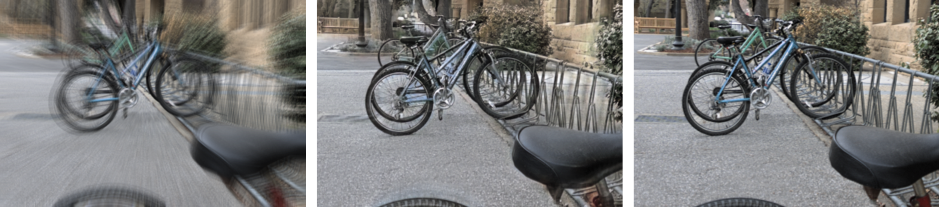}}
\end{minipage}
   \caption{An example of LF deblurring task. The three images are the blurred image, the sharp image recovered by our method, and the ground truth, respectively. Our method can get the deblurring results of all views through the network only once.}
\label{fig:deblur_example}
\end{figure}

In this paper, we use 4D function $L(u,v,x,y)\in \mathbb{R}^{U\times V\times X\times Y}$ 
to represent LFs, where $(u,v)$ and $(x,y)$ are the angular and spatial coordinates, respectively. If we fixed $(u,v)$, the Sub-Aperture Image (SAI) $I_{u,v}$ can be obtained. If we fixed $(x,y)$, the Micro-lens Images $M_{x,y}$ can be obtained. 
Different from the single image blur, the LF blur can be modeled as:
\begin{equation}
    L_{B}(u,v,x,y) =\int_{P^t_{u,v,x,y}}L^t_{S}(u,v,x,y)\, dt,
\end{equation}
where $L_{B}(u,v,x,y)$ denotes the blurred LFs, $L^t_{S}(u,v,x,y)$ denotes the sharp LFs, $P^t_{u,v,x,y}$ denotes the motion path of the pixel $(x,y)$ in view $(u,v)$ within the exposure time $t$. 
Compared with the single image, which has only spatial dimension $(x,y)$, the LFs have extra angular dimension $(u,v)$. Therefore, the blur phenomenon in LFs is more complex than the single image. Fig.~\ref{fig:deblur_example} is an example of LF deblurring task. On one hand, when the light field camera moves, the motion path of each view relative to the object is different, so the blur degree of each view is also different. On the other hand, similar to a single image, objects with different depths produce different degrees of blur, i.e., blur is more serious in pixels with small depth and less in pixels with large depth.

Single image methods~\cite{Zhang-et-al:dmphn,Zamir-et-al:mprnet} are possible ways to deal with LF deblurring task.
However, for the LF blur is affected by the views and depth, if the single image methods are directly applied to the LF deblurring task, i.e., dealing with each view independently, their results cannot maintain the LF structure. Since the LFs are collected regularly in the angular dimension, the LFs have the consistent structure, i.e., the correspondence between different views.
Furthermore, since the single image methods cannot effectively utilize the LF angular and depth information, their results are still blurred in some blur areas. Although the LFs with abundant angular information has greater potential to deal with this problem, LF deblurring methods~\cite{Srinivasan-et-al:lf2,Lee-et-al:lf3,Lumentut-et-al:lf5} are still in their infancy. Traditional methods~\cite{Srinivasan-et-al:lf2,Lee-et-al:lf3} are often time-consuming. Although the LF method~\cite{Lumentut-et-al:lf5} based on deep learning process images quickly, they destroy the LF consistent structure, because the LFs are divided into multiple groups in their network.
By analyzing the current research status of the single image and LF methods, we summarize three urgent problems in the LF deblurring task:
\textbf{1)} How to deal with the problem of different blur degrees in different views. \textbf{2)} How to solve the different blur degrees caused by different depths in the same view. \textbf{3)} How to maintain the LF consistency.

In this paper, based on the above problems we find and summarize, we propose a novel end-to-end learning-based framework. For the different blur degrees in different views, our method calculates the exclusive convolution kernels for different views to perceive and deal with the corresponding blur. For the pixels with different blur degrees from different depths, the mixed range between the sharp pixel we want to recover and the other view pixels is different in the corresponding micro-lens image. Therefore, we designed a depth perception view attention for micro-lens images to pick out the sharp pixel information, which is mixed in other view pixels. Then the sharp images are restored pixel by pixel based on implicit depth information. Besides, an angular position embedding is applied to depth perception view attention to maintain the LF consistency.

Our main contributions can be summarized:
\begin{itemize}
    \item We propose a novel view adaptive spatial convolution to deblur the different views, which automatically adapts to the blur degree of different views.
    \item We design a novel depth perception view attention to restore sharp images pixel by pixel, which deals with the different blurs caused by different depths. Besides, the angular position embedding is applied to maintain the LF consistency. 
    \item Quantitative and qualitative experimental results on synthetic and real images show that our method is superior to other state-of-art methods.
\end{itemize}

\section{Related Work}

This section will introduce the related works from single image and LF methods. Since the application scenarios of the blind methods are more widely than unblinded methods, 
we only analyze the blind deblurring method that the blur kernel is unknown.

\subsection{Single Image Deblurring Method}
In the past few decades, single image deblurring task has been studied extensively and remarkable achievements have been achieved. Traditional methods
remove image blur by various prior knowledge such as color~\cite{Joshi-et-al:s2}, gradient ~\cite{Chen-et-al:s3}, dark channel ~\cite{Pan-et-al:s4},and L0 regularization~\cite{Pan-et-al:s5}. 
However, these methods are often computationally complex and cannot achieve satisfactory results on complex real datasets.

~\cite{Sun-et-al:s7,Chakrabart-et-al:s9} are the  pioneers to solve the deblurring problem based on deep learning, and made remarkable progress. Following these pioneers, multi-scale convolution~\cite{Nah-et-al:s11,Suin-et-al:s14,Zhang-et-al:dmphn,Zamir-et-al:mprnet} and Recurrent Neural Networks~\cite{Zhang-et-al:s10,Park-et-al:s15} are applied to model the spatial variation blur kernel in dynamic scenes. Inspired by the success of Generative Adversarial Networks, ~\cite{Kupyn-et-al:s12,Kupyn-et-al:s13} generated the sharp images more in line with human visual perception.

Although the single image methods are very effective for processing single image deblurring, the LF blur is different from single image. The single image methods have some limitations in processing LFs: \textbf{1)} The single image methods do not use the abundant angular information in LFs, so that the depth information is missing during the deblurring processing. By contrast, our LF method makes full use of depth information by combining complementary information from all views in LFs, so that the blur boundaries in different depths can be better handled. \textbf{2)} Using the single image methods to deal with each view separately cannot maintain the LF angular structure very well, which will break the corresponding relationships between views in LFs. By contrast, we process all views at one time, which maintains the LF structure well.

\subsection{LF Deblurring Method}
The LFs belong to multi-images, but it is different from other multi-image types such as video and stereo images. Although these methods~\cite{Pan-et-al:m1,Zhang-et-al:m2,Wang-et-al:m3,Zhou-et-al:m4,Pan-et-al:m5,Zhong-et-al:m6} are excellent in dealing with blur in their field, we do not describe them in detail for their image types are different with us.

With the maturity of the commercialization of light field cameras, a variety of applications based on light fields have been brought. 
Although LFs have been widely studied in the fields of super-resolution~\cite{Wang-et-al:lfsr1,Jin-et-al:lfsr2} and depth estimation~\cite{Chen-et-al:lfde1,Zhou-et-al:lfde2}, there is little research on LF deblurring.
Dansereau et al.~\cite{Dansereau-et-al:lf1} first proposed the non-blind LF deblurring method based on the known camera motion. Srinivasan et al.~\cite{Srinivasan-et-al:lf2} are the first to model the 3-DOF camera motion and proposed a blind LF deblurring method on a 3-DOF dataset. 
Different from 3-DOF camera motion, 6-DOF methods ~\cite{Lumentut-et-al:lf4,Lee-et-al:lf3} deal with the LF deblurring task on the 6-DOF motion. However, the computational complexity of these traditional methods is too high, and the deblurring effect is seriously limited by the estimation of camera motion. 


The existing LF deblurring method ~\cite{Lumentut-et-al:lf5} based on deep learning does not make full use of the LF spatial and angular structure. Since the LFs are divided into multiple groups, the LF structure is destroyed. Different from them, the LFs are input as a whole in our method. Therefore, the restoration results of all views are obtained only by one calculation, which effectively maintains the LF structure.

\begin{figure}[!t]
\centerline{\includegraphics[width=1.0\linewidth]{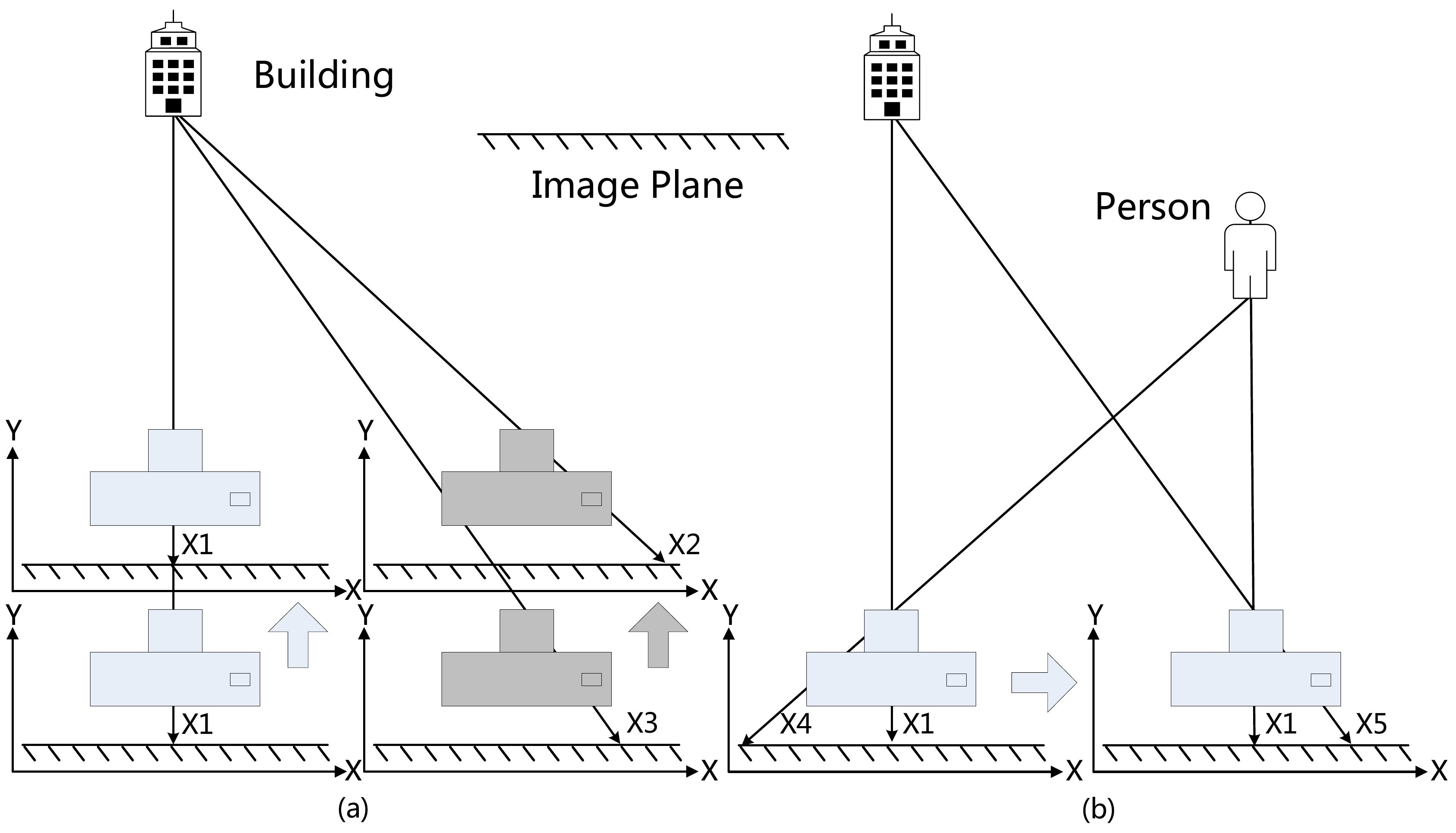}}
\caption{Schematic diagram of LF blurred image. Different colors represent different cameras, and arrows represent the direction of camera motion. }
\label{fig:Deblurex}
\end{figure}

\begin{figure*}[!t]
\centerline{\includegraphics[width=1.0\linewidth]{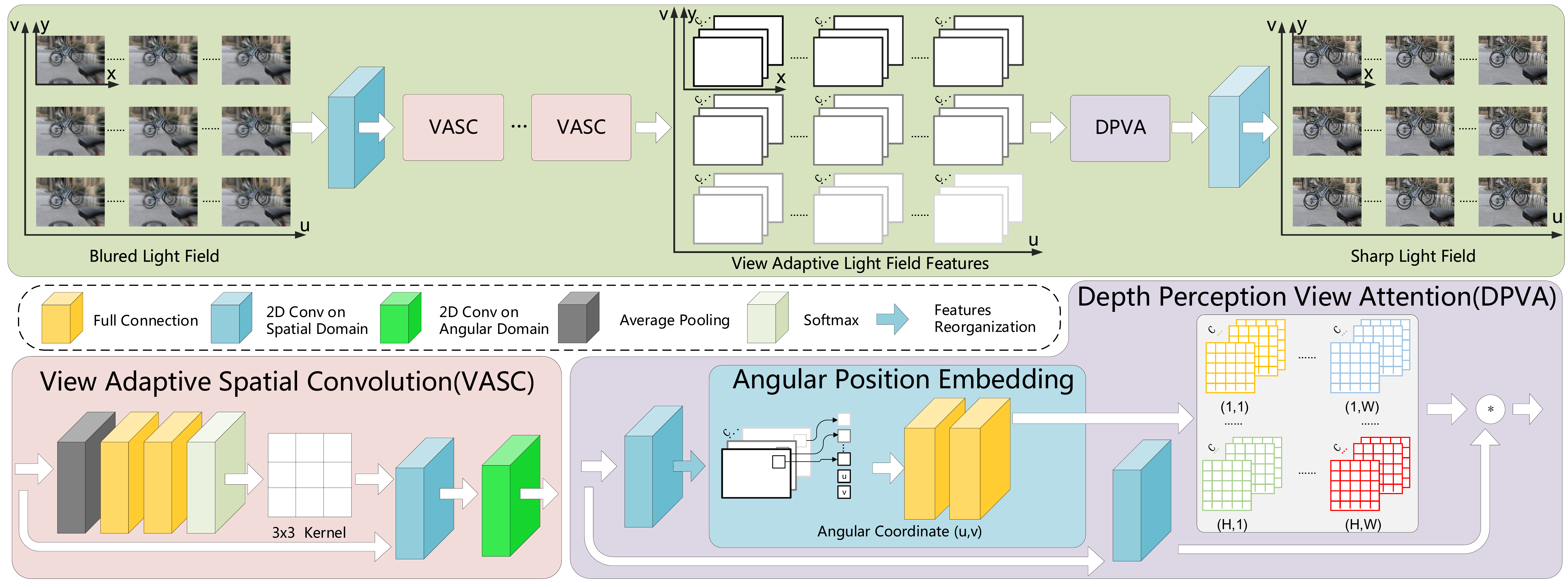}}
\caption{The network architecture of our model contains two main modules: the $VASC$ module and the $DPVA$ module. $VASC$ module calculates the corresponding convolution kernel for each view. It is repeated several times to get a more refined recovery result. $DPVA$ module fuses the weighted micro-lens image pixel by pixel by sensing the blur range caused by depth. To maintain the particularity of each view and the consistency of LFs, the $APE$ module is added to the $DPVA$ module.}
\label{fig:Deblur}
\end{figure*}

\section{Motivation}
\label{Mo}

Inspired by stereo deblurring scenes~\cite{Zhou-et-al:dava}, we found there are two characteristics of blurred LFs, which are useful for the LF deblurring task. On one hand, the blur degree of the same object is different in different views. On the other hand, the blur degree of different objects with various depths is also different. Fig.~\ref{fig:Deblurex} is a schematic diagram of blurred LFs to explain the two characteristics. For convenience, we only show two cameras in Fig.~\ref{fig:Deblurex}, which can be easily extended to the case of LF.

In Fig.~\ref{fig:Deblurex} (a), the building is on the same vertical line as one of the cameras and the light emitted from the building reaches the image plane after passing through the camera lens. We suppose the camera moves in a simple straight line along the Y-axis. As the two cameras move along the Y-axis, the position that the light of the building reaches on the left image plane $x1$ does not change, but on the right image plane, the position change from $x3$ to $x2$. For more complex rotational motion, this phenomenon will be more obvious. 
Therefore, the blur degree of the building in different views is different since the different relative motion between the different view cameras and the object.  
If we share the parameters of the convolution kernel for all views when designing the model, the model uses the same convolution kernel to deal with different blur degrees, which is unreasonable. 
Noting this, we design a novel view adaptive spatial convolution, which dynamically calculates the exclusive convolution kernel for each view without significantly increasing the parameters. The detail is introduced in Sec.~\ref{VASC}.

In Fig.~\ref{fig:Deblurex} (b), the person and the building have different depths. As the camera move along the X-axis, the position that the light of the building reaches on the image changes from $x1$ to $x5$, while the position that the light of the person reaches on the image changes from $x4$ to $x1$. As we can see from Fig.~\ref{fig:Deblurex} (b), $x5-x1$ is less than $x1-x4$, which means that when the camera moves, the blur degree of the objects with different depths is different. Generally, the blur degree of objects with a small depth is large, and the blur degree of objects with a large depth is small. 

To make full use of the LF angular information, we introduce the micro-lens image. In the micro-lens image, the sharp pixel we want to recover is mixed with other view pixels. For the different blur degrees brought by depth, the mixed range between the sharp pixel and other pixels is also different. If the blur degree of the sharp pixel is serious, the mixed range is large, otherwise, it is small. When restoring a sharp pixel, we should focus on the positions where the sharp pixel is mixed.
Therefore, we design a depth perception view attention to find the area where the sharp pixels are mixed. Then the sharp images are restored pixel by pixel by fusing the corresponding weighted micro-lens image features. The detail is introduced in Sec.~\ref{DPVA}.


\begin{table*}
\caption{Quantitative results (PSNR/SSIM/NCC/LMSE) on 3-DOF and 6-DOF datasets. $\uparrow$ denotes the higher value the better, $\downarrow$ is the opposite. Bold indicates the best one.}
\centering
\begin{tabular}{l|l|cccc||cccc}
\toprule
\multirow{2}[0]{*}{Type} &\multirow{2}[0]{*}{Method} &\multicolumn{4}{c||}{3-DOF}
&\multicolumn{4}{c}{6-DOF}\\
\cline{3-10}
& & PSNR $\uparrow$ & SSIM $\uparrow$& NCC $\uparrow$ &LMSE $\downarrow$& PSNR $\uparrow$ & SSIM $\uparrow$ & NCC $\uparrow$ &LMSE $\downarrow$\\
\hline
\multirow{2}[0]{*}{2D}&MPRNet~\cite{Zamir-et-al:mprnet} & 23.53 & 0.6989 &0.9009&0.0304 & 24.45 & 0.7835 & 0.9433 & 0.0153 \\
\cline{2-10}
&DMPHN~\cite{Zhang-et-al:dmphn}  & 25.61 & 0.7396 & 0.9291&0.0208 & 24.67 & 0.7673 &  0.9509& 0.0136\\
\hline
\multirow{5}[0]{*}{LF}&SAS~\cite{Yeung-et-al:sas}  & 25.58 & 0.8432 & 0.9451&0.0128 & 24.57 & 0.7698 &  0.9491& 0.0143\\
\cline{2-10}
&LFBMD~\cite{Srinivasan-et-al:lf2}  & 18.38 & 0.4720 &0.7546&0.0657 & 18.57 & 0.6997 & 0.9305 & 0.0169 \\
\cline{2-10}
&MEGNet~\cite{Zhang-et-al:mag}  & 26.36 & 0.8643 &0.9544&0.0111 & 24.81 & 0.7851 & 0.9519& 0.0139 \\
\cline{2-10}
&InterNet~\cite{Wang-et-al:inter} & 25.63 & 0.8547 &0.9474&0.0123 & 24.89 & \textbf{0.7897} &\textbf{0.9530} & 0.0139 \\
\cline{2-10}
&Ours  & \textbf{27.50} & \textbf{0.8695} & \textbf{0.9641} & \textbf{0.0096}& \textbf{25.06} & 0.7706 & 0.9527&\textbf{0.0132 }\\
\bottomrule
\end{tabular}

\label{tab:Quantitative resluts}
\end{table*}

\section{Our Method}

The overview architecture of our proposed method is shown in Fig.~\ref{fig:Deblur}. The blurred LFs are first fed into the View Adaptive Spatial Convolution ($VASC$) module, which dynamically calculates the exclusive convolution kernel for each view.
After that, the sharp LFs can be reconstructed pixel by pixel through the Depth Perception View Attention ($DPVA$) module, which calculates different attention weights for objects with different depths to pick out the sharp pixel information in the corresponding micro-lens images.
Besides, to maintain the consistency of LFs, the Angular Position Embedding ($APE$) module is introduced into $DPVA$ module.

\subsection{View Adaptive Spatial Convolution}
\label{VASC}
As mentioned in Sec.~\ref{Mo}, the blur degrees of the same object are different in different views. 
There are two intuitive ways to deal with it.
\textbf{1)} The model ignores the blur degrees difference of different views and shares the same convolution kernel for all views. \textbf{2)} The model simply assigns a convolution kernel to each view.
For the first way, if all views share the same convolution kernel, it is difficult for a model to perceive different blur degrees. The lack of the ability to perceive the blur degrees will lead the model to deal with the average blur degree of all views. For the second way, if simply assign a convolution kernel to each view when building the model, the number of model parameters will increase exponentially. For example, assuming that the parameter required to process one view is $C_{in} \times C_{out} \times k \times k$, where $C_{in}$, $C_{out}$, and $k$ denotes the number of input channels, the number of output channels, and the convolution kernel size, respectively.
Therefore, the parameter required to process $N_{view}$ views according to the above method is $N_{view} \times C_{in} \times C_{out} \times k \times k$.
In order to enable the model to perceive the difference of blur degrees between different views without significantly increasing the number of parameters,
$VASC$ module is cleverly designed, which is shown in Fig.~\ref{fig:Deblur}. 

Firstly, the blurred LFs $L_{B}(u,v,x,y)$ are input into the $VASC$ module in the form of SAIs. Then the exclusive convolution kernel for each view are generated.
Take one view as an example, the SAI features are fed into one average pooling and two full connections to generate the corresponding view adaptive convolution kernel $W_{u,v}\in \mathbb{R}^{k\times k}$.
Then our method extracts the view adaptive spatial features based on spatial convolution by $W_{u,v}$.
Finally, the all view adaptive features $F_{VA}\in \mathbb{R}^{U\times V\times X\times Y\times C}$ are fused through angular convolution.
\begin{equation}
    F_{VA}(u,v,x,y) =H_{VASC}(L_{B}(u,v,x,y)).
\end{equation}
The number of $VASC$ module is $C_{in} \times C_{K} + C_{K}\times C_{K} +C_{K}\times C_{in}\times C_{out} \times k \times k$, where $C_{K}$ denotes the number of nodes in the two full connection layers. In this paper, $C_{K}$ =4 < $N_{view}$=25, which denotes that our model has fewer parameters. For more details, please refer the codes in the supplementary materials.

By using the plug and play $VASC$ module iteratively, our model not only effectively perceives the difference of blur degrees in different views but also does not significantly increase the number of parameters.


\subsection{Depth Perception View Attention}
\label{DPVA}
As mentioned in Sec.~\ref{Mo}, the blur degree of objects with different depths in the same view is different. To make full use of the LF angular information, we introduce the micro-lens image, where the the sharp pixel we want to recover is mixed with other pixels from other views.
Since the different blur degrees, the mixed range in micro-lens images is also different.
Therefore, we designed the $DPVA$ module to find the mixed area for the different blur degrees pixels and then pick out the sharp pixel information based on implicit depth information, which is shown in Fig.~\ref{fig:Deblur}.


Specifically, the view adaptive features $F_{VA}$ from the $VASC$ module are fed into two branches. 
In the lower branch, our method generates the view exclusive LF features $F^{ve}_{u,v}\in \mathbb{R}^{ X\times Y\times UVC}$ for all views, through a 2D convolution layer that the output channel number is $U\times V$ times larger than the input channel number.


In the upper branch, the view adaptive features $F_{VA}$ are fed into another 2D convolution layer to change channel dimension from $C$ to $UV$, then $U\times V$ depth perception features $F^{dp}_{u,v}\in \mathbb{R}^{X\times Y\times UV}$ can be obtained. 
To maintain interaction between all views,
we reorganize all depth perception features as $F^{ndp}_{u,v}\in \mathbb{R}^{X\times Y\times UV}$.
Taking the (u,v) view as an example, the $uv$-th channel dimension of all depth perception features $F^{dp}_{u,v}$ is connected as the feature $F^{ndp}_{u,v}$. 
To further maintain the consistency of LFs, the $APE$ module is applied to the $DPVA$ module to generate the depth perception features with angular position embedding $F^{adp}_{u,v}$, which is introduced in Sec.~\ref{ape}.
After that, the $F^{adp}_{u,v}$ are fed into several full connection layers to get Depth Perception View Attention Weight $W^{dp}_{u,v} \in \mathbb{R}^{ X\times Y\times UVC}$, which denotes the whether the position in the micro-lens image has information that needs to be restored.
Then, the sharp features $F^{sharp}_{u,v}\in \mathbb{R}^{ X\times Y\times C}$ can be obtained by the view exclusive LF features $F^{ve}_{u,v}$ and the View Attention Weight $W^{dp}_{u,v}$:
\begin{equation}
\label{eq_att}
    F^{sharp}_{u,v}(x,y,c) =\sum_{\hat{u}\hat{v}}F^{ve}_{u,v}(x,y,\hat{u}*\hat{v}*c)* W^{dp}_{u,v}(x,y,\hat{u}*\hat{v}*c).
\end{equation}

Finally, the sharp LFs $L_{S}$ can be generated from the sharp features $F^{sharp}$ through a 2D convolution. 
\begin{equation}
    L_{S}(u,v,x,y) =H_{DPVA}(F_{VA}(u,v,x,y)).
\end{equation}

Our method takes the whole LFs as the input and uses and maintains the LF spatial and angular structure as much as possible. On the contrary, the past deep learning LF methods only input part of the LFs into the network at a time. They not only cannot make full use of the LF structure but also need multiple calculations to obtain the results of all views.

\subsection{Angular Position Embedding}
\label{ape}
Without the angular position embedding, the model cannot know which view is being processed. The lack of positional embedding causes the model to tend to handle the average of all views, which destroys the LF structure.
Therefore, the $APE$ module is necessary for the $DPVA$ module, which can be seen in Fig.~\ref{fig:Deblur}. Since the angular dimension of the LFs is fixed when acquiring the LFs, we directly use the absolute position as position embedding. Taking the (u,v) view as an example, the angular coordinates (u,v) is concatenated with $F^{ndp}_{u,v}$ to get $F^{adp}_{u,v}\in \mathbb{R}^{X\times Y\times (UV+2)}$, which changes the channel dimension from $U\times V$ to $U\times V + 2$.
 
\begin{equation}
    F^{adp}_{u,v}(x,y)=H_{APE}(F^{ndp}_{u,v}(x,y),(u,v)).
\end{equation}



\section{Experiments}
In this section, we first introduce the dataset and training details. Then the proposed method is compared with other single image methods and LF methods on synthetic and real images quantitatively and qualitatively. Next, the ablation study is taken to prove the effectiveness of the designed modules. Finally, the limitation of our method is discussed.

\subsection{Training Details and Datasets}

\subsubsection{Dataset.}
There are two main kinds of datasets (3-DOF and 6-DOF) in the LF deblurring task. In the 3-DOF dataset, the camera shifts in X, Y, and Z directions. Besides the 3-DOF motions, the camera rotates in X, Y, and Z directions in the 6-DOF dataset. In this paper, we trained models on 3-DOF and 6-DOF datasets separately, since the camera motions are different.
We picked $200$ LFs from Stanford Lytro ~\cite{Raj-et-al:lytro} as a 3-DOF training dataset and $40$ LFs as a 3-DOF test dataset. The simulation method of blurred images is consistent with that of ~\cite{Srinivasan-et-al:lf2}. Since~\cite{Lumentut-et-al:lf4} only published 40 images of 6-DOF datasets. We selected the 3 images used in~\cite{Lumentut-et-al:lf4} as the 6-DOF test dataset and others as the training dataset.

\subsubsection{Training Details.}
In this paper, we crop the training LFs into $5\times5\times64\times64\times3$ patches. Image flipping and image rotation are applied as data augmentation. We use RGB images as input and output. The angular size of input and output are both $5\times5$. The number of $VASC$ modules is $8$.
The proposed model is optimized using the Adam~\cite{Kingma-et-al:adam} algorithm with a batch size of $4$. The initial learning rate is set as $1e-3$ in the first $200$ epochs and is then divided by $10$ every $100$ epochs. We implement the model with the PyTorch framework and the training process roughly takes $2$ days for the 3-DOF dataset and $1$ days for the 6-DOF dataset with $4$ Intel(R) Xeon(R) CPU E5-2683 v3 @ 2.00GHz with a Titan XP GPU.

\subsubsection{Loss Function.}
Our method uses the L1 Loss as supervision:
\begin{equation}
\begin{footnotesize}
  Loss_{L1}= \left\|L_{S}-L_{gt} \right\|_1,
\end{footnotesize}
\end{equation} 
where $L_{gt}$ denotes the ground truth.

\begin{figure*}[!t]
\small
\begin{minipage}{.19\linewidth}
  \centerline{Origin}
\end{minipage}
\begin{minipage}{.19\linewidth}
  \centerline{Ground Truth}
\end{minipage}
\begin{minipage}{.19\linewidth}
  \centerline{DMPHN~\cite{Zhang-et-al:dmphn}}
\end{minipage}
\begin{minipage}{.19\linewidth}
  \centerline{MPRNet~\cite{Zamir-et-al:mprnet}}
\end{minipage}
\begin{minipage}{.19\linewidth}
  \centerline{Ours}
\end{minipage}

\begin{minipage}{1\linewidth}
   \centerline{\includegraphics[width=1\linewidth]{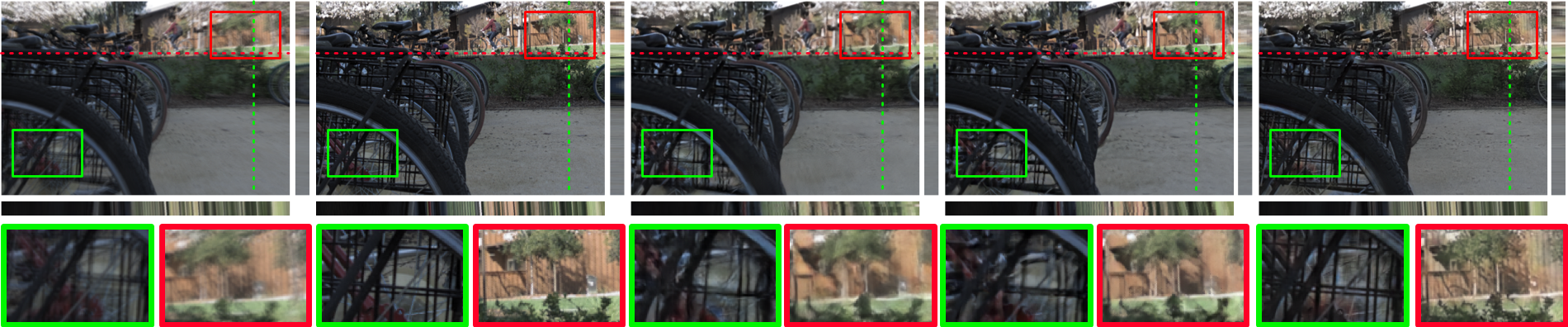}}
\end{minipage}

\begin{minipage}{.19\linewidth}
  \centerline{}
\end{minipage}
\begin{minipage}{.19\linewidth}
  \centerline{PSNR/SSIM}
\end{minipage}
\begin{minipage}{.19\linewidth}
  \centerline{24.80/0.7564}
\end{minipage}
\begin{minipage}{.19\linewidth}
  \centerline{23.75/0.7053}
\end{minipage}
\begin{minipage}{.19\linewidth}
  \centerline{27.49/0.8886}
\end{minipage}

\begin{minipage}{1\linewidth}
   \centerline{\includegraphics[width=1\linewidth]{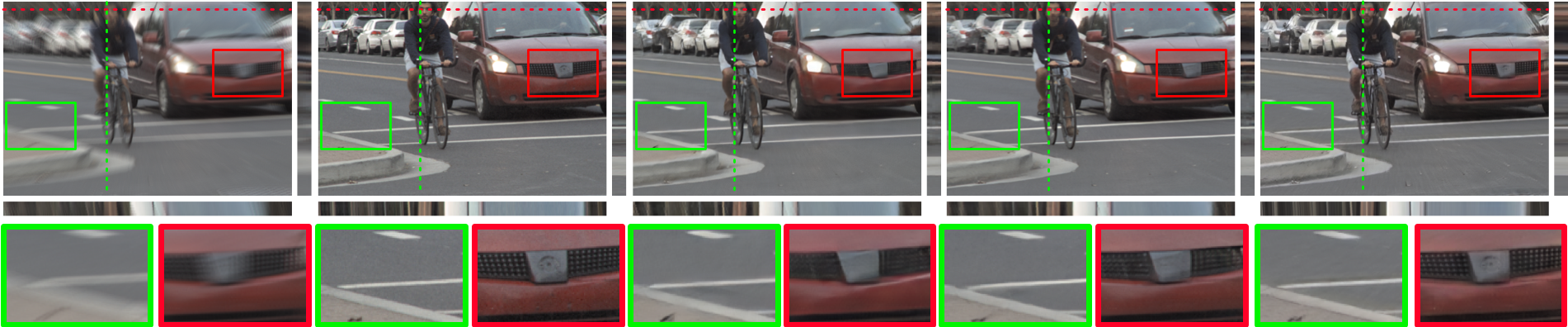}}
\end{minipage}

\begin{minipage}{.19\linewidth}
  \centerline{}
\end{minipage}
\begin{minipage}{.19\linewidth}
  \centerline{PSNR/SSIM}
\end{minipage}
\begin{minipage}{.19\linewidth}
  \centerline{29.38/0.8637}
\end{minipage}
\begin{minipage}{.19\linewidth}
  \centerline{29.81/0.8898}
\end{minipage}
\begin{minipage}{.19\linewidth}
  \centerline{31.04/0.9032}
\end{minipage}

\begin{minipage}{1\linewidth}
   \centerline{\includegraphics[width=1\linewidth]{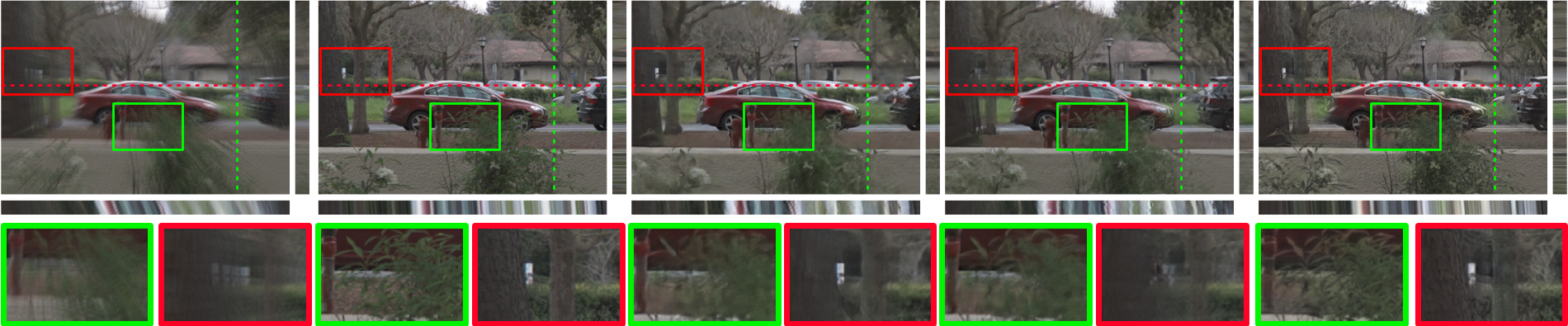}}
\end{minipage}

\begin{minipage}{.19\linewidth}
  \centerline{}
\end{minipage}
\begin{minipage}{.19\linewidth}
  \centerline{PSNR/SSIM}
\end{minipage}
\begin{minipage}{.19\linewidth}
  \centerline{28.63/0.7459}
\end{minipage}
\begin{minipage}{.19\linewidth}
  \centerline{24.29/0.6221}
\end{minipage}
\begin{minipage}{.19\linewidth}
  \centerline{30.19/0.9047}
\end{minipage}

\begin{minipage}{1\linewidth}
   \centerline{\includegraphics[width=1\linewidth]{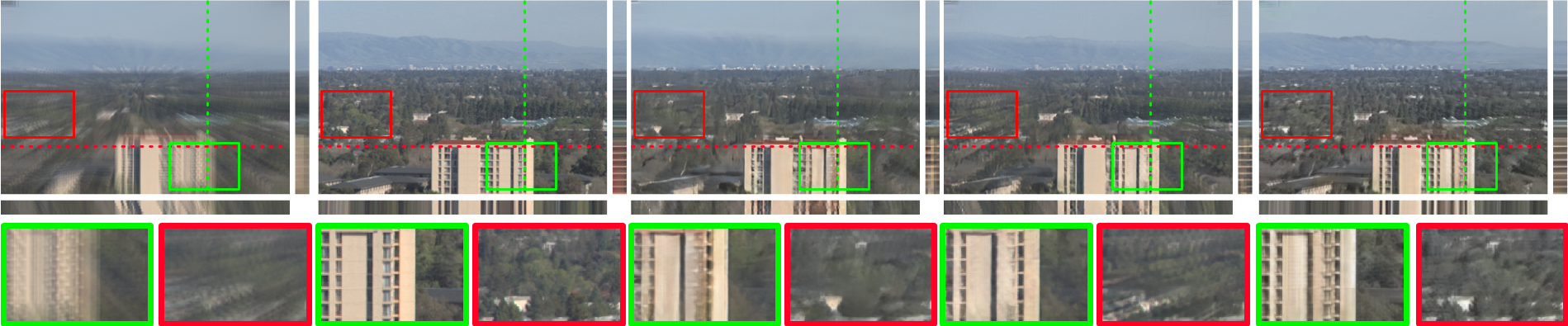}}
\end{minipage}

\begin{minipage}{.19\linewidth}
  \centerline{}
\end{minipage}
\begin{minipage}{.19\linewidth}
  \centerline{PSNR/SSIM}
\end{minipage}
\begin{minipage}{.19\linewidth}
  \centerline{26.92/0.7826}
\end{minipage}
\begin{minipage}{.19\linewidth}
  \centerline{27.22/0.8004}
\end{minipage}
\begin{minipage}{.19\linewidth}
  \centerline{30.05/0.8987}
\end{minipage}

\caption{Deburring results on test datasets with single image methods. To compare the performance of various methods to maintain the LF structure, we show the horizontal and vertical EPIs. In EPIs, if the line segment is missing or bent, it indicates that the consistency of the LF structure is destroyed. It can be seen that our method is much better than other single image methods in maintaining the consistency of the LF structure.}
\label{fig:Comparison_single}
\end{figure*}

\begin{figure*}[!t]
\small
\begin{minipage}{.11\linewidth}
  \centerline{Origin}
\end{minipage}
\begin{minipage}{.14\linewidth}
  \centerline{Ground Truth}
\end{minipage}
\begin{minipage}{.14\linewidth}
  \centerline{LFBMD~\cite{Srinivasan-et-al:lf2}}
\end{minipage}
\begin{minipage}{.14\linewidth}
  \centerline{SAS~\cite{Yeung-et-al:sas}}
\end{minipage}
\begin{minipage}{.14\linewidth}
  \centerline{InterNet~\cite{Wang-et-al:inter}}
\end{minipage}
\begin{minipage}{.14\linewidth}
  \centerline{MEGNet~\cite{Zhang-et-al:mag}}
\end{minipage}
\begin{minipage}{.14\linewidth}
  \centerline{Ours}
\end{minipage}

\begin{minipage}{1\linewidth}
   \centerline{\includegraphics[width=1\linewidth]{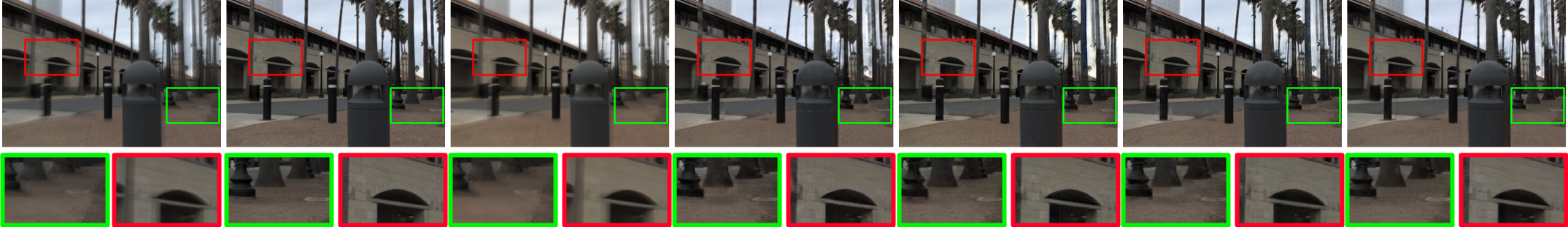}}
\end{minipage}

\begin{minipage}{.11\linewidth}
  \centerline{}
\end{minipage}
\begin{minipage}{.14\linewidth}
  \centerline{PSNR/SSIM}
\end{minipage}
\begin{minipage}{.14\linewidth}
  \centerline{15.93/0.5272}
\end{minipage}
\begin{minipage}{.14\linewidth}
  \centerline{26.11/0.8797}
\end{minipage}
\begin{minipage}{.14\linewidth}
  \centerline{26.26/0.8922}
\end{minipage}
\begin{minipage}{.14\linewidth}
  \centerline{26.75/0.8851}
\end{minipage}
\begin{minipage}{.14\linewidth}
  \centerline{27.21/0.9054}
\end{minipage}

\begin{minipage}{1\linewidth}
   \centerline{\includegraphics[width=1\linewidth]{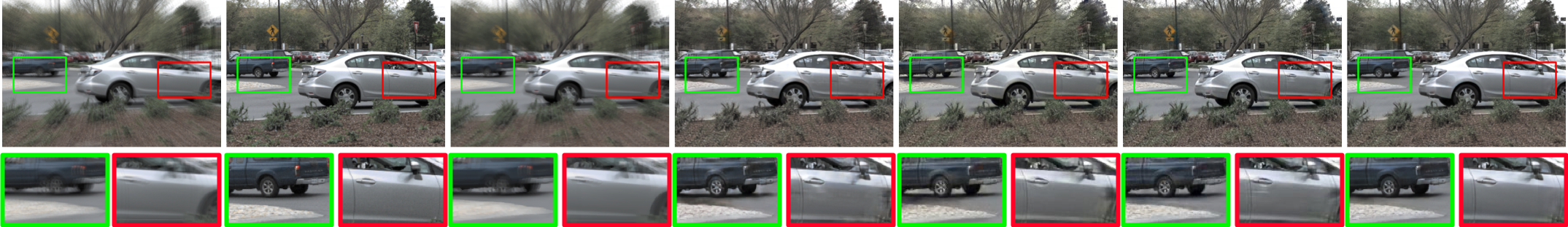}}
\end{minipage}

\begin{minipage}{.11\linewidth}
  \centerline{}
\end{minipage}
\begin{minipage}{.14\linewidth}
  \centerline{PSNR/SSIM}
\end{minipage}
\begin{minipage}{.14\linewidth}
  \centerline{18.73/0.4111}
\end{minipage}
\begin{minipage}{.14\linewidth}
  \centerline{26.49/0.8777}
\end{minipage}
\begin{minipage}{.14\linewidth}
  \centerline{25.75/0.8846}
\end{minipage}
\begin{minipage}{.14\linewidth}
  \centerline{27.14/0.0892}
\end{minipage}
\begin{minipage}{.14\linewidth}
  \centerline{28.92/0.9102}
\end{minipage}

\begin{minipage}{1\linewidth}
   \centerline{\includegraphics[width=1\linewidth]{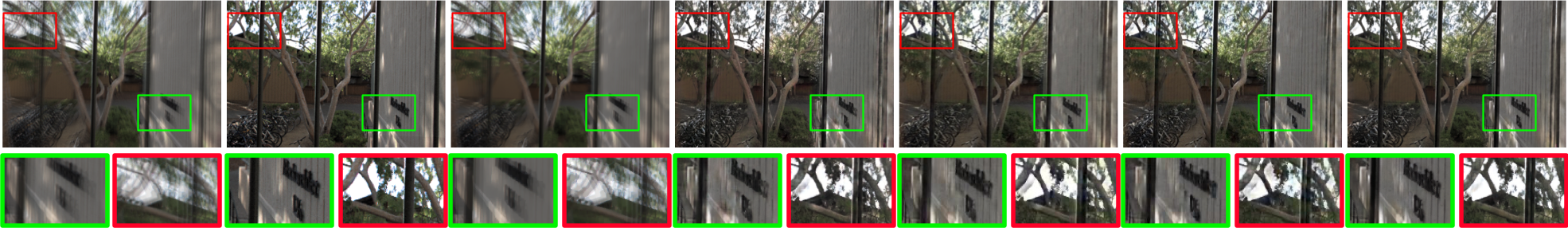}}
\end{minipage}

\begin{minipage}{.14\linewidth}
  \centerline{}
\end{minipage}
\begin{minipage}{.11\linewidth}
  \centerline{PSNR/SSIM}
\end{minipage}
\begin{minipage}{.14\linewidth}
  \centerline{16.54/0.4076}
\end{minipage}
\begin{minipage}{.14\linewidth}
  \centerline{23.91/0.8455}
\end{minipage}
\begin{minipage}{.14\linewidth}
  \centerline{24.17/0.8595}
\end{minipage}
\begin{minipage}{.14\linewidth}
  \centerline{24.69/0.8681}
\end{minipage}
\begin{minipage}{.14\linewidth}
  \centerline{25.51/0.8693}
\end{minipage}

\begin{minipage}{1\linewidth}
   \centerline{\includegraphics[width=1\linewidth]{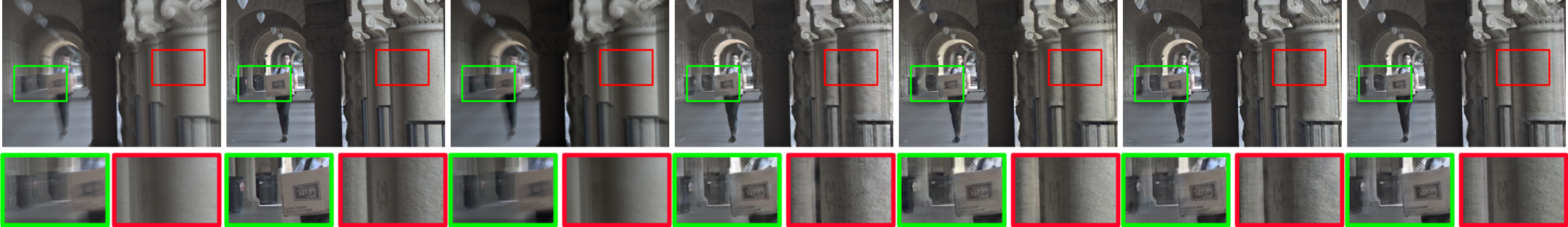}}
\end{minipage}

\begin{minipage}{.11\linewidth}
  \centerline{}
\end{minipage}
\begin{minipage}{.14\linewidth}
  \centerline{PSNR/SSIM}
\end{minipage}
\begin{minipage}{.14\linewidth}
  \centerline{17.33/0.5188}
\end{minipage}
\begin{minipage}{.14\linewidth}
  \centerline{26.45/0.8261}
\end{minipage}
\begin{minipage}{.14\linewidth}
  \centerline{25.46/0.8399}
\end{minipage}
\begin{minipage}{.14\linewidth}
  \centerline{26.95/0.8516}
\end{minipage}
\begin{minipage}{.14\linewidth}
  \centerline{30.04/0.8657}
\end{minipage}

   \caption{Deburring results on test datasets with LF methods.
   }
\label{fig:Comparison_LF}
\end{figure*}

\subsection{Quantitative Comparison}
We compared two single image methods DMPHN~\cite{Zhang-et-al:dmphn} and MPRNet~\cite{Zamir-et-al:mprnet} with excellent results in the field of single image deblurring.
Although there are some deblurring methods based on LFs, most of them do not publish codes and datasets~\cite{Lee-et-al:lf3,Lumentut-et-al:lf4,Lumentut-et-al:lf5}. Fortunately, LFBMD~\cite{Srinivasan-et-al:lf2} opened their codes so that we can easily compare with them.
Since the LF super-resolution methods also explore the LF spatial and angular information, which is conducive to deblurring, we modify three LF super-resolution methods SAS~\cite{Yeung-et-al:sas}, InterNet~\cite{Wang-et-al:inter}, and MEGNet~\cite{Zhang-et-al:mag} for LF deblurring task.

Tab.~\ref{tab:Quantitative resluts} is the quantitative results of 3-DOF and 6-DOF test datasets. The PSNR, SSIM, NCC, and LMSE are used for the evaluation index. The higher the values of PSNR, SSIM, and NCC denote the better results. LMSE is the opposite. Tab.~\ref{tab:Quantitative resluts} shows our method is much better than other single image and LF methods in all evaluation indexes on 3-DOF test datasets.
On the 6DOF test dataset, our method also achieved the best results on PSNR and LMSE, and other evaluation indexes also achieved comparable results.
Because the single image methods DMPHN~\cite{Zhang-et-al:dmphn} and MPRNet~\cite{Zamir-et-al:mprnet} process each view alone, they cannot effectively use the angular information of the LFs, i.e., they cannot obtain complementary information from other views. On the contrary, our LF method takes the whole LFs as the input, which fully explores and makes use of the angular and spatial information, so it achieves a better restoration effect.
For other LF methods LFBMD~\cite{Srinivasan-et-al:lf2}, SAS~\cite{Yeung-et-al:sas},    InterNet~\cite{Wang-et-al:inter}, and MEGNet~\cite{Zhang-et-al:mag}, they also use the angular information of LFs, but they do not consider that the blur degree is affected by the views and depth, so they cannot perceive the difference of views and depth well. Different from them, our method designs the $VASC$ module and $DPVA$ module to deal with the different blur degrees in different views and depths, which makes our method more effective.

\begin{figure*}[!t]
\begin{minipage}{.16\linewidth}
  \centerline{Origin}
\end{minipage}
\begin{minipage}{.16\linewidth}
  \centerline{Ground Truth}
\end{minipage}
\begin{minipage}{.16\linewidth}
  \centerline{w/o VASC}
\end{minipage}
\begin{minipage}{.14\linewidth}
  \centerline{w/o DPVA}
\end{minipage}
\begin{minipage}{.16\linewidth}
  \centerline{w/o APE}
\end{minipage}
\begin{minipage}{.16\linewidth}
  \centerline{Final}
\end{minipage}
\begin{minipage}{.02\linewidth}
  \centerline{}
\end{minipage}
\centerline{\includegraphics[width=1.0\linewidth]{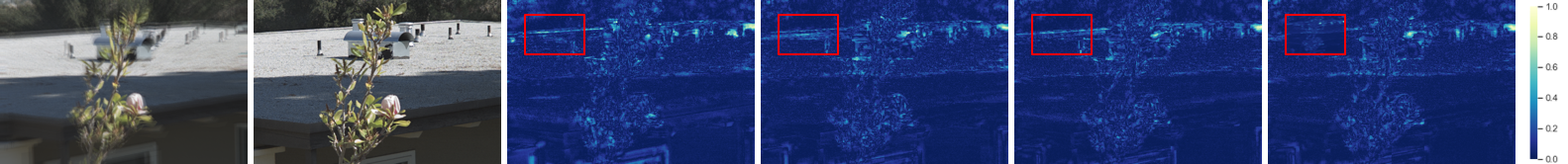}}
\centerline{\includegraphics[width=1.0\linewidth]{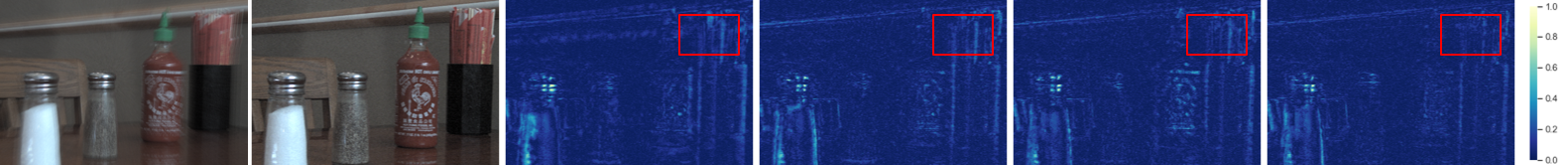}}
\caption{The error map of ablation study. The error map shows the difference of the predicted result and the ground truth. The closer to blue, the smaller the difference, and the closer to yellow, the greater the difference.}
\label{fig:ab_error}
\end{figure*}

Tab.~\ref{tab:time and parameter} shows the average time of processing a $5 \times 5 \times 500 \times 336$ LFs by various methods and the parameters of various models.
For the single image methods DMPHN~\cite{Zhang-et-al:dmphn} and MPRNet~\cite{Zamir-et-al:mprnet}, time is the sum of all views.
LFBMD~\cite{Srinivasan-et-al:lf2} is a traditional method that the parameter is not displayed. LFBMD~\cite{Srinivasan-et-al:lf2} needs to predict the motion trajectory of the camera first and needs a complex optimization process, so it is very time-consuming to process an LF, which greatly limits the application scenarios. 
The time consumption of LF methods based on deep learning is similar to that of single image methods.
However, the parameters of the LF methods are much smaller than those based on single image. Specifically, the parameters of our method are 30 times less than MPRNet~\cite{Zamir-et-al:mprnet} and 33 times less than DMPHN~\cite{Zhang-et-al:dmphn}.


\begin{table}
\caption{Parameters and the running time for one $5 \times 5 \times 500 \times 336$ LFs. The single image methods processes the each view independently, and finally sums all the time.}
\centering
\begin{tabular}{l|l|cc}
\toprule
Type &Method & Time/s &Param/M\\
\hline
\multirow{2}[0]{*}{2D}&MRPNet~\cite{Zamir-et-al:mprnet} & 2.31 & 19\\
\cline{2-4}
&DMPHN~\cite{Zhang-et-al:dmphn} &3.37 & 21\\
\hline
\multirow{5}[0]{*}{LF}&SAS~\cite{Yeung-et-al:sas} & 2.62 &0.63\\
\cline{2-4}
&LFBMD~\cite{Srinivasan-et-al:lf2} &$>$3600 & /\\
\cline{2-4}
&MEGNet~\cite{Zhang-et-al:mag} & 2.12 & 0.42\\
\cline{2-4}
&InterNet~\cite{Wang-et-al:inter} & 3.51 &8\\
\cline{2-4}
&Ours  & 3.21 & 0.63\\
\bottomrule
\end{tabular}
\label{tab:time and parameter}
\end{table}

\subsection{Qualitative Comparison}
Fig.~\ref{fig:Comparison_single} is the qualitative comparison with single image methods. We show the result of the central view and enlarge some areas to make the comparison more obvious. In Fig.~\ref{fig:Comparison_single}, our results are more clear than the results of the single image methods, and the details are better preserved. To better show the performance of various methods for maintaining the LF structure, we introduce the Epipolar Plane Images (EPIs). In EPIs, the slope of the line segment represents the disparity between two adjacent views. If the LF structure is maintained well, the line segment will not bend or break. 
Since the single image methods process each view separately, there are bending and fracture in the EPIs of the single image methods in Fig.~\ref{fig:Comparison_single}. On the contrary, our method takes the LFs as a whole and introduce the $APE$ module to embed angular position, the line segment in our EPIs is a straight line, which fully proves that our method can maintain the structural consistency of LFs.

Fig.~\ref{fig:Comparison_LF} is the qualitative comparison with LF methods on the test dataset.
Fig.~\ref{fig:Comparison_LF} shows that the result of LFBMD is still blurred and is easy to cause color deviation. This is because LFBMD needs to manually set multiple hyperparameters, which limits the application scenario. When the scene becomes complex, the setting of hyperparameters will not be accurate enough, which will easily lead to the deterioration. 
Different from it, the LF methods based on deep learning do not need to set complex hyperparameters, since they learn automatically from a large amount of data.
Because these LF methods SAS~\cite{Yeung-et-al:sas}, InterNet~\cite{Wang-et-al:inter}, and MEGNet~\cite{Zhang-et-al:mag} are not specially designed for deblurring tasks, they cannot perceive the different blur degrees in different views and depths. Compared with our methods, the results of these methods have some obvious artifacts, which can be seen in the enlarged image.
\begin{figure}[!t]
\begin{minipage}{.22\linewidth}
  \centerline{Origin}
\end{minipage}
\begin{minipage}{.25\linewidth}
  \centerline{MPRNet~\cite{Zamir-et-al:mprnet}}
\end{minipage}
\begin{minipage}{.25\linewidth}
  \centerline{DMPHN~\cite{Zhang-et-al:dmphn}}
\end{minipage}
\begin{minipage}{.25\linewidth}
  \centerline{Our}
\end{minipage}
\centerline{\includegraphics[width=1.0\linewidth]{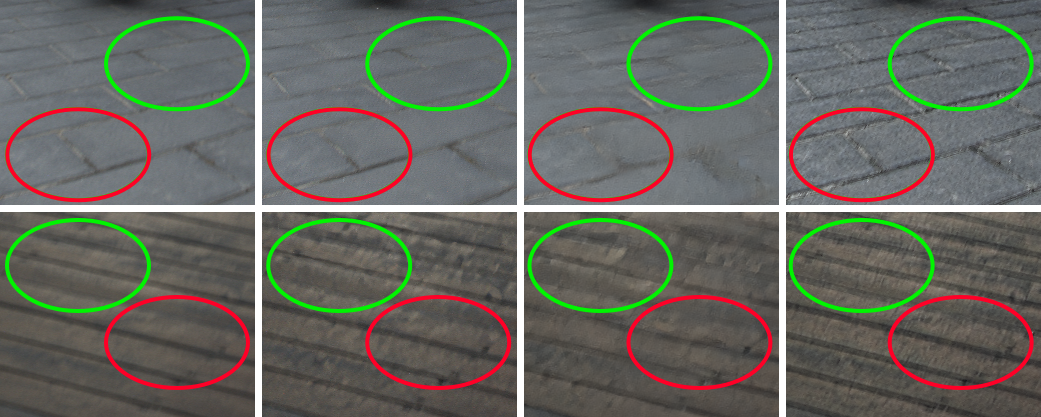}}
\caption{Deburring results on real scenes.
}
\label{fig:real}
\end{figure}
Fig.~\ref{fig:real} is the qualitative comparison of real
scenes. Since there is no published real LF blur dataset, we use a Lytro Illum LF camera to obtain the real blur scene for visual comparison. Our results are sharper than the single image methods, which shows our method is more robust in real scenes.

\subsection{Ablation Study}

\begin{table}
\caption{Ablation studies results (PSNR/SSIM/NCC/LMSE) on the 3-DOF test dataset.
}
\centering
\begin{tabular}{l|cccc}
\toprule
Method &PSNR $\uparrow$& SSIM $\uparrow$& NCC  $\uparrow$ &LMSE $\downarrow$\\
\hline
w/o VASC & 25.27 & 0.8346 &0.9414&0.0136\\
\hline
w/o DPVA & 26.98 & 0.8601&0.9598&0.0106 \\
\hline
w/o APE  &26.78&  0.8585&0.9567 &0.0107\\
\hline
Final & \textbf{27.50} & \textbf{0.8695} & \textbf{0.9641} & \textbf{0.0096} \\
\bottomrule
\end{tabular}
\label{tab:Ablation}
\end{table}

To verify the effectiveness of our method, we did some ablation studies on the 3-DOF test dataset.
To ensure the fairness of the ablation study, we keep the parameters of the models almost the same. The experimental results are shown in Tab~\ref{tab:Ablation}. 
First, we replace the $VASC$ module with the ordinary convolution (w/o $VASC$) which is shared by all views. The significant decline in the evaluation indexes shows that ordinary convolution cannot learn the prior knowledge that the blur degree of different views is different. On the contrary, the $VASC$ module calculates an exclusive convolution kernel for each view effectively.
Then the $DPVA$ module is removed completely (w/o $DPVA$) to verify it can generate the different view attention weights for regions with different depths. The experimental results in Tab.~\ref{tab:Ablation} show that the $DPVA$ module improves the performance by fusing the angular information in the micro-lens image.
Without the $APE$ module (w/o $APE$), the performance decreases significantly. This is because the $APE$ module helps to maintain the LF structural consistency. Besides, if there is no $APE$ module, the $DPVA$ module cannot realize what information needs to obtain from the micro-lens image for the specific view, which explains why the performance without $APE$ module is worse than that without $DPVA$ module.
Fig.~\ref{fig:ab_error} shows the error map of ablation study. We can see that no matter $VASC$ module, $DPVA$ module, or $APE$ module is missing, there will be more error areas, which shows the effectiveness of the structure we designed again. Our method fully explores the LF blur characteristics by the $VASC$ and $DPVA$ module. Therefore, our results are sharper and the details are restored more perfectly.
\subsection{Limitation}
\begin{figure}[!t]
\begin{minipage}{.30\linewidth}
  \centerline{Origin}
\end{minipage}
\begin{minipage}{.33\linewidth}
  \centerline{Our}
\end{minipage}
\begin{minipage}{.33\linewidth}
  \centerline{InterNet~\cite{Wang-et-al:inter}}
\end{minipage}
\centerline{\includegraphics[width=1.0\linewidth]{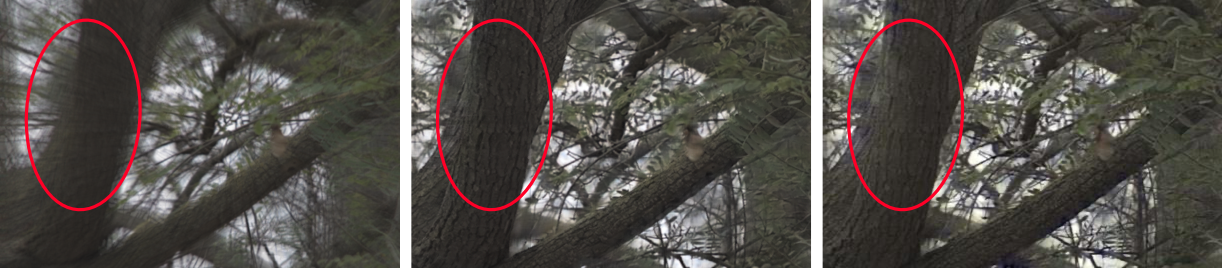}}
\begin{minipage}{.30\linewidth}
  \centerline{PSNR/SSIM}
\end{minipage}
\begin{minipage}{.33\linewidth}
  \centerline{23.62/0.8282}
\end{minipage}
\begin{minipage}{.33\linewidth}
  \centerline{22.73/0.8257}
\end{minipage}
\begin{minipage}{.30\linewidth}
  \centerline{Ground Truth}
\end{minipage}
\begin{minipage}{.33\linewidth}
  \centerline{MPRNet~\cite{Zamir-et-al:mprnet}}
\end{minipage}
\begin{minipage}{.33\linewidth}
  \centerline{DMPHN~\cite{Zhang-et-al:dmphn} }
\end{minipage}
\centerline{\includegraphics[width=1.0\linewidth]{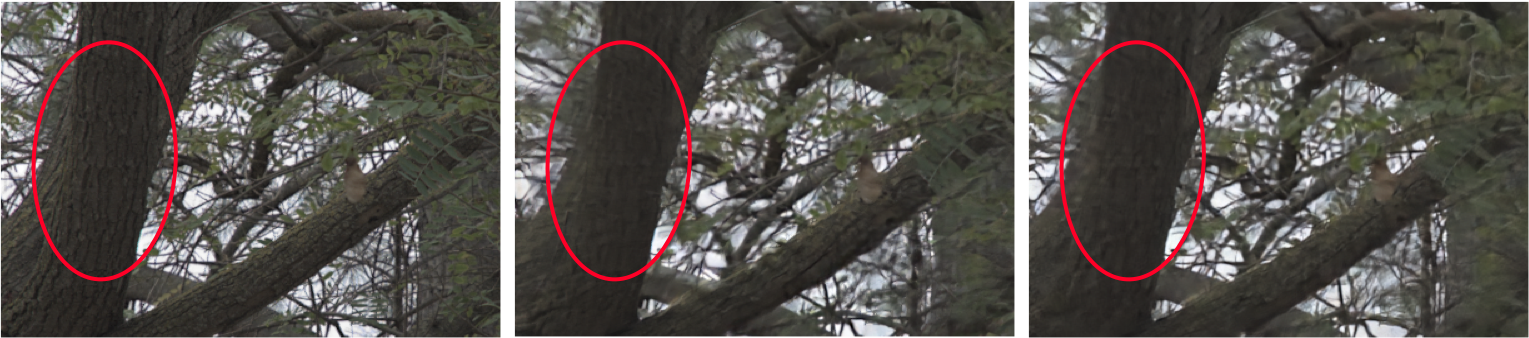}}
\begin{minipage}{.30\linewidth}
  \centerline{PSNR/SSIM}
\end{minipage}
\begin{minipage}{.33\linewidth}
  \centerline{20.01/0.5963}
\end{minipage}
\begin{minipage}{.33\linewidth}
  \centerline{23.72/0.7478}
\end{minipage}
\caption{A failure case. The first/second line is the result of two LF/single image methods. In this scence, the performance of 
LF methods degenerates to the single image methods.
}
\label{fig:failure}
\end{figure}
Fig.~\ref{fig:failure} is a failure case of all single image and LF methods. In this scene, blur is very serious in all views. Our LF method cannot effectively obtain useful information from other views to restore the sharp image. Therefore, the ability of deblurring degenerates to the single image deblurring methods. Nevertheless, the detail of our result is still better than the single image methods, especially the restoration of trunk texture.

\section{Conclusions}
In this paper, we proposed a novel LF deblurring method based on deep learning. 
First, the $VASC$ module is designed to effectively deal with different blur degrees in different views by calculating an exclusive convolution kernel for each view. Besides, we designed a $DPVA$ module to deal with the different blur degrees in different depths. 
Through this module, the sharp results are recovered by the weighted micro-lens image based on implicit depth information.
Furthermore, the $APE$ module is also applied to maintain the overall consistency of the LFs. 
Our LF method takes the whole LFs as a whole and utilizes the complementary angular information, which makes it to better maintain the structural consistency of the LFs than the single image methods.
Quantitative and qualitative experimental results show that our method is much better than other single image and LF deblurring methods.

\bibliographystyle{ACM-Reference-Format}
\bibliography{sample-base}

\appendix









\end{document}